\documentclass[pdflatex,sn-mathphys]{sn-jnl}

\jyear{2021}%

\theoremstyle{thmstyleone}%

\theoremstyle{thmstyletwo}%

\theoremstyle{thmstylethree}%

\begin{document}

\title{Scale-Invariant Object Detection by Adaptive Convolution with Unified Global-Local Context}

\author[1]{\fnm{Amrita} \sur{Singh}}\email{as998@snu.edu.in}

\author*[1]{\fnm{Snehasis} \sur{Mukherjee}}\email{snehasis.mukherjee@snu.edu.in}

\affil*[1]{\orgname{Shiv Nadar Institution of Eminence}, \orgaddress{\city{Delhi NCR}, \country{India}}}

\abstract{Dense features are important for detecting differnet scale objects in images. Unfortunately, despite of the remarkable efficacy of the CNN models in multi-scale object detection, CNN models often fail to detect different scale objects in images due to using similar type of CNN features . Atrous convolution address this issue by applying sparse kernels. However, sparse kernels often can lose the multi-scale detection efficacy of the CNN model. In this paper, we propose an object detection model using a Switchable Atrous Convolutional Network (SAC-Net) based on the efficientDet model. A fixed atrous rate limits the performance of the CNN models in the convolutional layers. To overcome this limitation, we introduce a switchable mechanism that allows for dynamically adjusting the atrous rate during the forward pass. The proposed SAC-Net encapsulates the benefits of both low-level and high-level features to achieve improved performance on multi-scale object detection tasks, without losing the dense features. Further, we apply a depth-wise switchable atrous rate to the proposed network, to improve the scale-invariant features. Finally, we apply global context on the proposed model. Our extensive experiments on benchmark datasets demonstrate that the proposed SAC-Net outperforms the state-of-the-art models by a significant margin in terms of accuracy. The codes are available at }\url{https://github.com/anAmrita/SAC_Net/tree/master}.


\maketitle

\section{Introduction}\label{sec1}
The detection of multiple objects in images is a classical problem in the field of computer vision. The goal of object detection is to find a set of given objects in the image. Object detection is an active area of interest among researchers in computer vision, as object detection serves as the basis for several high-level computer vision tasks in applications such as robot vision, surveillance, autonomous driving, content-based image retrieval, human-computer interaction, and many more. Despite being a well-studied problem during the past few decades, object detection still remains an unsolved problem due to the various challenges associated with the problem, such as variation in scale, variation in viewpoint, context, lighting conditions, and many more. The variety of objects and the intra-class variations within the object category, add to the challenges in detecting objects.

The task of object detection can be of two types: object instance detection and generic object detection. The goal of object instance detection is to detect all instances of a particular object, such as a particular breed of dog, the Eiffel tower, etc. Whereas, the goal of generic object detection is to detect instances of the given set of categories of objects such as cats, dogs, cars, bicycles, buildings, etc., in the image. Most of the object detection methods found in the literature are of the first category. Comparatively less attention was given to generic object detection. This study proposes a deep learning-based method for generic object detection.

With the introduction of deep learning-based techniques (especially CNNs), the efficacy of object detectors has enhanced significantly during the last few years. The recent CNN-based object detectors can efficiently detect objects in challenging datasets such as COCO \cite{coco}. However, detecting objects with smaller appearances still remains an unsolved problem, despite a few attempts to detect smaller objects \cite{imavis_survey}.

The recent deep learning-based object detection approaches can be categorized into two major classes: two-stage process and one-stage process \cite{ijcv_survey}. The two-stage or region-based object detection approaches generate the region proposals (class-independent) in the first stage. In the second stage, the CNN features are extracted from the region proposals, and fed into a classifier for classification \cite{fasterrcnn,cascrcnn,fpn,sppnet,rpn_cvpr16,dcn}. The one-stage or unified object detection approaches propose a single feed-forward CNN to directly predict the object category, without generating the region proposals \cite{efficientdet,detectors,tood,yolo}. In general, two-stage approaches provide much better accuracy compared to the one-stage approaches, however, two-stage approaches are computationally heavy. Although there are efforts found in the literature to obtain the efficacy of the two-stage detectors by a one-stage detector, by using the same backbone CNN architecture for both the stages of the detection process \cite{eccv_mimicnet}. However, computational complexity still remains a problem. One-stage object detectors are gaining popularity because of their simpler architecture and less training time \cite{efficientdet,detectors,yolo}.

The introduction of the You-Only-Look-Once (YOLO) family of object detectors has been a revolutionary step towards one-stage object detection in real-time \cite{yolo}. However, one-stage object detectors often suffer from memory overhead during training, because of a huge number of model parameters. Tan et al. reduced the parameter overhead of the one-stage object detector significantly, by efficient use of the EfficientNet architecture \cite{efficientdet}. Qiao et al. proposed switchable atrous convolution on the backbone CNN model of the object detector, to automatically switch the atrous rates \cite{detectors}. Atrous convolution-based backbone model improved the model performance. However, \cite{detectors} applied the atrous convolution on a feature pyramid network, which complicated the detector model.

In this study, we applied the concept of switchable atrous convolution (motivated by \cite{detectors}), on a lighter model called EfficientDet \cite{efficientdet}, to reduce the GPU overhead and FLOPs for the proposed method. Further, we apply the switchable atrous rates depthwise, at the backbone model, to enhance the effect of atrous convolution in the model accuracy. Finally, we apply global context before and after the depthwise convolution layers, to make the model scale invariant. The contributions of this study is summarized as follows:

 \begin{itemize}
    \item In this proposed work using an adaptive atrous Conv layer with different atrous rates, capable of adapting with varying depths of the architecture.
    \item We apply Global context before and after the depthwise convolution layers, to make the proposed method scale-invariant.
    \item We apply the depthwise atrous convolution alongwith global context, on a lightweight EfficientDet model, to enhance the model performance in terms of model parameters.
\end{itemize}
   
Next, we provide a survey of the literature on object detection.

\section{Related Works}\label{sec2}
Generic object detection is an active area of interest among researchers in the computer vision area \cite{ijcv_survey}. The recent deep learning based approaches for object detection can be categorized into two classes: two-stage approaches, where region proposal generation is an intermediate stage towards object detection, and one-stage approaches, which directly detect the objects in images.

\subsection{Two-stage Object Detection}
Two-stage object detection techniques consist of two stages: first detecting the region proposals from the image, followed by detecting objects at the regions of interest \cite{fasterrcnn,cascrcnn,fpn,sppnet,rpn_cvpr16,dcn}. Two-stage object detection became popular because of the huge success of the RCNN family of object detectors \cite{rcnn,fastrcnn, fasterrcnn, cascrcnn}. The R-CNN proposed the concept of identifying the region of interest (ROI) from the image, using a VGG Net-based CNN architecture, followed by categorization of the ROI into an object class by another CNN \cite{rcnn}. The ROIs are again proposed from the test images, which are categorized by the trained classifier. R-CNN shown state-of-the-art performance in terms of accuracy, however, it suffers from huge training time (because of the two stages of CNN layers). He et al. refined this approach by replacing the set of CNNs for ROI generation, with a single CNN on the whole image \cite{sppnet}. The ROIs are extracted by Spatial Pyramid Pooling (SPP), reducing the training time for the first stage. Fast R-CNN proposed the same concept of using a single CNN for ROI extraction, followed by SPP, however, they have refined the region proposals by sending the region proposals through some fully connected (FC) layers \cite{fastrcnn}. Further, Fast R-CNN applies a single-layer SPP network for ROI pooling from the region proposal CNN. This reduces the execution time during testing.

Ren et al. proposed Faster R-CNN by proposing Region Proposal Network (RPN) for proposal generation \cite{fasterrcnn}. Unlike Fast R-CNN, where ROI pooling is applied directly on the CNN feature, Faster R-CNN applies RPN to extract the ROI. Cascaded R-CNN followed the concept of Faster R-CNN, where bounding box regression was performed in cascaded regions \cite{cascrcnn}. Lin et al. proposed Feature Pyramid Network (FPN) within the Faster R-CNN framework, to achieve state-of-the-art accuracy in reduced time \cite{fpn}. Srivastava et al. trained the region pooling network of the Fast R-CNN model using hard examples \cite{rpn_cvpr16}. Dai et al. proposed deformable convolution network during the ROI pooling process in the proposal generator architecture \cite{dcn}.

Two-stage object detectors provide better accuracy compared to one-stage object detectors, due to the focus on generating the object region as the first step \cite{ijcv_survey}. However, the models for two-stage detectors are much more complex compared to the one-stage detectors due to the use of two CNNs in a series, resulting in high computation costs. Despite a few attempts to reduce the execution time of a two-stage object detection process. training time and memory overhead are the major problems in two-stage object detectors. Recently researchers are focusing more on developing one-stage object detectors, with improved accuracy.

\subsection{One-stage Object Detection}
One-stage object detectors aim to provide an end-to-end network for object detection, directly from the image. One-stage object detectors are becoming famous because of their simple network structure and lesser computational overhead compared to two-stage object detectors. One-stage object detectors gained popularity following the success of the YOLO family of object detectors \cite{yolo}, providing good detection accuracy in almost real-time.

YOLO considers the task of object detection as a regression problem, where the image pixels are mapped into spatially separated bounding boxes \cite{yolo}. YOLO uses a small set of candidate regions, to directly obtain the object regions. YOLO divides the image into different grids, each predicting the class probabilities and bounding boxes. Despite being fast, YOLO often fails in detecting varying scale of objects in the image, because of the grid formation while candidate region generation \cite{ijcv_survey}.

Several later versions of YOLO were proposed to improve the performance of YOLO object detector \cite{yolov2,yolov3,yolov4}. YOLOv2 replaces the backbone GoogleNet architecture with a much simpler DarkNet-19 architecture, alongwith batch normalization \cite{yolov2}. Further, YOLOv2 removed the fully connected (FC) layer from the YOLO model, making it even simpler. YOLOv3 further revised the YOLOv2 model, by replacing the backbone DarkNet-19 with DarkNet-53 model \cite{yolov3}. YOLOv4 introduced some more features including residual connections, addressing overfitting during training \cite{yolov4}. YOLOv5 and higher versions use transformer architecture \cite{yolov5}. Li et al. applied multi-feature fusion on the YOLOv3 network, to deal with varying scales of objects \cite{paa2}. The main problem in YOLO family of object detectors remains the same: overlooking varying scales of objects due to the grid-based object detection approach. Efforts are made for conditional adaptation of object scales, however, Gilg et al. have shown that such conditional adaptations are biased towards object sizes \cite{gilg_wacv23}.

A few efforts were made to address the problem of detecting smaller objects by the one-stage detectors \cite{tood,detectors,efficientdet}. Feng et al. framed the one-stage object detection task as a combination of two tasks: object localization and object classification \cite{tood}. Further, they proposed a multi-task learning strategy, to make the two tasks collaborate. Such a multi-task learning strategy has a high GPU computational overhead. Tan et al. proposed EfficientDet, a bi-directional feature pyramid network (Bi-FPN) based on the EfficientNet as the backbone, to perform a multi-scale feature fusion, to help detect objects in different scales, in less computational overhead \cite{efficientdet}. Qiao et al. proposed a recursive feature pyramid network, alongwith switchable atrous convolution, to convolve the features in different atrous rates \cite{detectors}. The switchable atrous convolution mechanism enables better detection of objects with different scales.

There are efforts in the literature, to combine the benefits of region proposal generation (better accuracy) and the single-stage structure (simplicity), by applying a shared backbone for the two stages in the detection process \cite{eccv_mimicnet}. However, the complexity of the network still remains a problem. Efforts have been made by applying a deep gradient network, to deal with camouflaged objects \cite{mir}. Liu et al. proposed a feature enhancement module, to further improve the performance of the FPN, to detect smaller objects \cite{paa1}. Hou et al. \cite{tvc24_app} combined the concept of the region proposal with YOLOv5, to detect the defects on a surface.

Recent deep learning advancements have shown significant progress in medical image analysis focusing on object detection \cite{dai2024deep,qian2024drac,nazir2020off}. Prior systems, such as Dai et al. \cite{dai2024deep} developed by Google's DeepMind, have achieved high accuracy in diabetic retinopathy (DR) detection from fundus images. Qian et al. \cite{qian2024drac} provided a benchmark for AI algorithms in DR diagnosis using ultra-wide OCTA images. By standardizing performance metrics, they aimed to enhance AI models for early DR detection and monitoring, fostering advancements in AI-driven healthcare solutions. Dai et al. \cite{dai2021deep} aimed at real-time image quality assessment, lesion detection, and grading of fundus images, before detection. It uses interpretable AI models to identify key DR indicators, enhancing clinical efficiency and enabling timely interventions to prevent vision loss in diabetic patients. Qin et al. \cite{qin2024urbanevolver} optimized urban layout regeneration by integrating function awareness into the design process, enhancing traffic flow, public accessibility, and spatial organization to better meet the needs of urban inhabitants.

Efforts have been made to apply a deformable attention module in a transformer for object detection \cite{lin2021eapt,xie2021bagfn}. Unsupervised domain adaptation is employed by selecting the most informative and representative target samples for adaptation by Zhang et.al \cite{zhu2024clustering} for enhanced object detection. Chen et al. \cite{chen2023mngnas} proposed a Neural Architecture Search (NAS) method to find a suitable student model for object recognition tasks, where the teacher model is trained for classification. Efforts are made to apply aesthetic rules \cite{jiang2022photohelper}.

Sheng et al. \cite{sheng2021improving} proposed an incremental learning procedure for the detection and tracking of objects in videos. Li et al. \cite{li2021automatic} proposed a multimodal cascaded convolutional neural network (MCCNN) integrating subnetworks such as YOLOv4 \cite{yolov4}, and Faster-RCNN \cite{fasterrcnn}. Efforts are made to automatically colorize objects in the images, for better detection \cite{cheng2015deep}. Fuzzy subsystems were employed by Guo et al. \cite{guo2019multiview} for multiview object detection. Sheng et al. \cite{sheng2018intrinsic} shown that a proper categorization of shading in an image helps in handling abrupt changes in illumination. Chen et al. \cite{chen2018outdoor} emphasized information within the image itself, without additional scene requirements.

Recently, a few different approaches have been found in the literature for one-stage object detection \cite{adverse,icme22,distillation,aploss,attention,tvc24}. Agrawal et al. \cite{tvc24} proposed geometric features for object detection, where occlusion prior is introduced for handling occlusion. Chen et al. proposed adversarial training for object detection, where AdvProp was used for classification \cite{adverse}. Jiang et al. could effectively reduce the computational overhead by enhancing the interaction between the localization and classification tasks using a gating head \cite{icme22}. Efforts were made to apply knowledge distillation for the localization of object regions for the detection \cite{distillation}. Xu et al. applied average precision loss in the detection network, to get enhanced performance \cite{aploss}. Recently, attention-based transformer architectures are also being used for object detection \cite{attention}.

Motivated by the recent success of the feature pyramid networks (FPN) in object detection \cite{detectors,efficientdet}, with reduced computational overhead, we apply an FPN network following \cite{detectors}. Further, to enhance the performance of the proposed model in detecting varying scale of objects, we applied switchable atrous convolution on the FPN following \cite{efficientdet}. Unlike \cite{efficientdet}, where switchable atrous rates are applied on the whole architecture, the proposed model applies depthwise atrous convolution, to emphasize objects appearing in different scales. Finally, we apply global average pooling, to further improve the proposed model. Next, we illustrate the proposed method in detail.

\section{Proposed Method}\label{sec3}
The proposed model consists of three major components:
(i) Depthwise switchable atrous Conv layer with different atrous rates, (ii) Global context before and after the depthwise convolution layers, and (iii) Depthwise atrous convolution along with global context, on a lightweight EfficientDet model \cite{efficientdet}, to enhance the model performance in terms of model parameters. However, before going to discuss the three components, we illustrate the backbone EfficientDet model.

\subsection{EfficientDet: The backbone model}
We use EfficientDet \cite{efficientdet} as the backbone model in the proposed method. EfficientDet is a state-of-the-art object detection architecture that balances accuracy and efficiency (in terms of model parameters) for object recognition in real-world scenarios. The Backbone of the EfficientDet model is the EfficientNet \cite{efficientnet} model. The EfficientNet model consists of several building blocks, including a stem block, multiple blocks of repeating CNN building blocks, and a head block.

1. Stem Block: The stem block consists of a series of convolutional and pooling layers that reduce the spatial resolution of the input image and extract initial features.

2. MBConv layers: MBConv layers consists of some repeating blocks comprising multiple sub-layers, including a depthwise separable convolution layer, a pointwise convolution layer, and a skip connection. These blocks are repeated numerous times to form a deep network that captures increasingly complex features from the input image.

The depthwise convolutional layer applies a separate convolutional filter to each input channel. This allows the network to independently learn spatial features in each channel without mixing them. The main benefit of this layer is that it reduces the number of parameters and computation required in the network while also improving the efficiency of the network.

The pointwise convolutional layer applies a $1 \times 1$ convolutional filter to the output of the depthwise convolutional layer. This operation helps to combine the spatial features learned by the depthwise convolutional layer across channels and can also reduce the number of output channels. The benefit of this layer is that it allows the network to learn more complex features by combining spatial features across channels while also reducing the computational cost.

3. Head Block: The head block consists of a series of fully connected and global average pooling layers that generate the final predictions based on the features extracted by the repeating blocks.

In addition to the backbone, EfficientDet further includes a set of auxiliary layers that help improve the model's accuracy. These layers include a bi-directional feature pyramid network (BiFPN) network, which combines multi-level features from the backbone to generate a high-quality feature map for object detection, and a class/box network, which predicts the class and location of objects in the image. The
BiFPN is a key component of the EfficientDet object detection model that enables efficient feature fusion across different resolutions and scales. BiFPN consists of a series of repeated blocks, each containing a set of lateral connections that combine features from adjacent scales, followed by a top-down and a bottom-up path for feature fusion. The lateral connections help to propagate features across scales, while the top-down and bottom-up paths enable efficient feature fusion by aggregating features from multiple scales.

In traditional feature pyramid networks, features are passed only in a top-down manner, which can lead to information loss and inefficient feature propagation. BiFPN addresses this issue by using a bidirectional flow of information, where features are passed both in a top-down and bottom-up manner. The top-down path in BiFPN starts with the highest resolution features and gradually reduces the resolution by downsampling. Conversely, the bottom-up path starts with the lowest resolution features and gradually increases the resolution by upsampling. By combining information from both paths, BiFPN is able to fuse features across different scales and resolutions in an efficient manner.

We apply atrous convolution operation at the MBConv layer of the EfficientDet architecture to combine the global and local feature information extracted from the image.

\subsection{Atrous convolution layer}
Atrous convolution is a type of convolution operation that uses a larger filter than the input data and includes more context information from surrounding pixels, allowing the network to learn more complex features from the input data \cite{detectors}. Due to the ability of atrous convolution to extract minute contextual information from the images, an atrous convolution-based model helps in finding objects appearing in different scales, even when the scale of the object is too small.

\begin{figure}
\vspace{-1cm}
     \centering
     \includegraphics[height=6cm,width=0.8
    \textwidth]{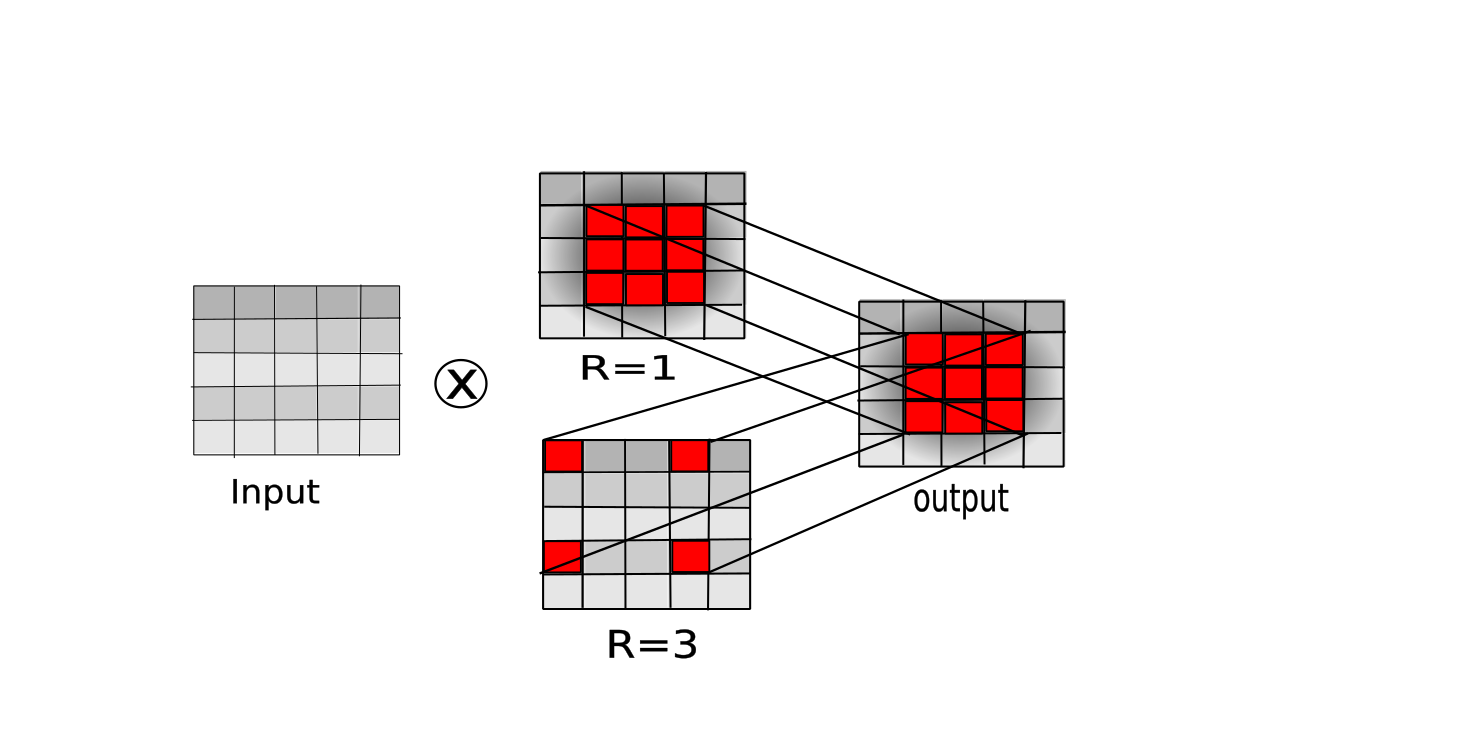}
     \caption{shows an input image convolved with different atrous rate convolution filters. R is the Atrous rate.}
     \label{convolving two filters}
\end{figure}

The formula for calculating the kernel dimension $k_d$ of an atrous convolutional layer is as follows.
\begin{equation} \label{eq1}
k_d = 1 + (a_r \times (k_s - 1))
\end{equation}
where $a_r$ is the Atrous rate, and $k_s$ is the kernel dimension of the CNN layers of the original EfficientDet architecture. Thus the kernel dimension for the CNN layer is adjusted after applying the atrous convolution operation. In order to deal with the revised dimension of the feature vector obtained from the atrous layers, 0-padding is applied on the feature vector as shown below:
\begin{equation} \label{eq2}
P_d = ((i_s - 1) \times s_t + k_d - i_s) / 2
\end{equation}
where $P_d$ is the updated feature vector after padding, $i_s$ the input size, and $s_t$ the stride. This padding is used for the input function only for the rank three convolution filter to obtain the same dimensional output. Fig. \ref{convolving two filters} shows the two filters with atrous rates 1 and 3, used in the proposed approach. The two atrous filters are convolving to give an output of the same dimensions. 

Further, in order to extract the global features from the image, we apply a global context block before and after the MBConv layer of the EfficientDet \cite{efficientdet} architecture.

\subsection{Global context block}
Global context refers to the ability of a neural network to capture information about the entire input data rather than just local features \cite{detectors}. Global pooling is one way to incorporate global context into a neural network. Global pooling allows the network to capture information about the overall structure of the input data.

First, the input feature $x$ is padded with a reflection padding of 2 pixels on each side using the torch.nn.functional.pad function. Reflection padding mirrors the values of the input tensor along the edges, creating a reflection of the input that can be used to provide context for features near the edge of the input. This is a common way to incorporate context information in CNNs.

After padding, the resulting tensor is passed through a 2D average pooling operation using the torch.nn.functional.avg\_pool2d function with a kernel size of 5x5, a stride of 1, and a padding of 0. This operation reduces the spatial dimensions of the tensor and computes the average value of each feature map across the entire spatial domain of the tensor.

In order to further leverage an efficient combination of global and local features to enable the proposed detector to detect objects of varying scales, we apply the switchable atrous convolution layer depthwise in multiple channels.

\subsection{Depthwise switchable atrous Conv layer with the different atrous rate (DSAC) }\label{subsec3}
\begin{figure}
     \centering
     \includegraphics[height=6cm,width=0.9
    \textwidth]{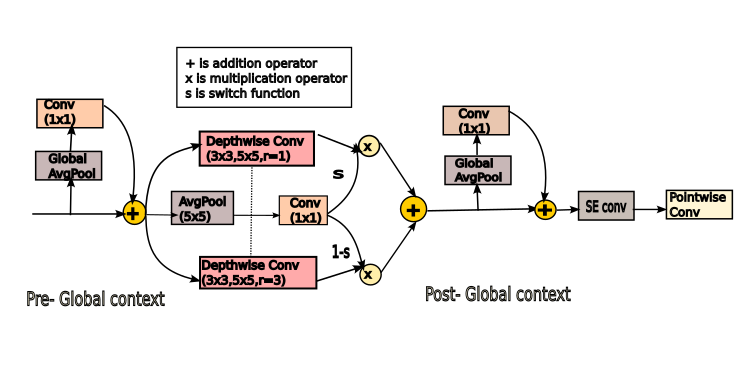}
     \vspace{-1cm}
     \caption{Depthwise switchable atrous Conv layer with different atrous(DSAC). We convert each $3\times3$ convolutional layer in the baseline Efficientnet to DSAC, which gradually alternates the atrous rates used for convolutional computation. Two global context modules add image-level information to the features.}
     \label{fig:2}
 \end{figure}

The depthwise switchable atrous Conv layer (DSAC) is introduced with different atrous rates in the proposed method. The proposed DSAC is influenced by \cite{detectors}. However, unlike \cite{detectors}, we applied the atrous convolution operation depthwise, instead of applying it on the backbone model, as done in \cite{detectors}. The proposed DSAC consists of three main components: two global context modules and one switchable atrous convolution layer. The two global context modules are added before and after the MBConv layer. Fig \ref{fig:2} presents the overall architecture of the proposed DSAC. MB conv has a depthwise Conv(DConv(x, w,1)) layer, a squeeze excitation layer(SE(x)), and a pointwise Conv layer (PConv(x)). In this study, we substitute depthwise separable convolution with depthwise switchable atrous Conv layer with different atrous rates.  We denote the depthwise convolutional operation with weight $w$ and atrous rate $r$ that takes $x$ as its input and outputs $y$ as $y = DConv(x, w, r)$. Thus, we can transform each depthwise Conv layer to DSAC in the following manner:
\begin{equation} \label{eq3}
 DConv(x, w,1)\longrightarrow S(x) . DConv(x, w,1) + (1-S(x) ) . DConv(x,w,r).
\end{equation}

The equation (\ref{eq3}) describes the operation of a DSAC with different atrous rates, which is a modified version of the depthwise separable convolution (DConv) layer. DSAC includes a switch function $(S(x))$ that dynamically selects either a DConv layer with an atrous rate of 1 or a DConv layer with a non-trivial atrous rate $r$ for each input feature map $x$. The operation of the DSAC involves passing the input feature map $x$ through a DConv layer with atrous rate 1, denoted as $DConv(x, w, 1)$, then using the switch function $S(x)$ to generate a binary mask that decides which operation to perform for each location in the feature map. The binary mask is then used to select either the output of $DConv(x, w, 1)$ or the output of $DConv(x, w, r)$ for each location in the feature map, and the selected feature maps are combined using element-wise addition to produce the final output feature map.

DSAC with different atrous rates is used to capture multi-scale features that are beneficial for object detection tasks. The switch function is implemented as an average pooling layer with a $5\times5$ kernel followed by a $1\times1$ convolutional layer, with $r$ set to 3 in the experiments. The switch function's weights are learned during training, and they define each feature map's contribution to the fusion process based on the object scale. The performances of DSAC with and without global context modules are shown in the ablation study, with global context increasing the performance of the detection.

With the DSAC consisting of the switchable atrous convolution layer and two global context blocks covering it, the image-level global information is extracted efficiently. In addition to the DSAC module, we further enhance the proposed global feature by extracting the global information at the feature level, using a Depthwise atrous convolution with a pointwise switchable Conv layer (DAPSC).

\subsection{Depthwise atrous with pointwise switchable Conv layer (DAPSC) }\label{subsec4}
\begin{figure}
     \centering
     \includegraphics[height=4cm,width=0.9
    \textwidth]{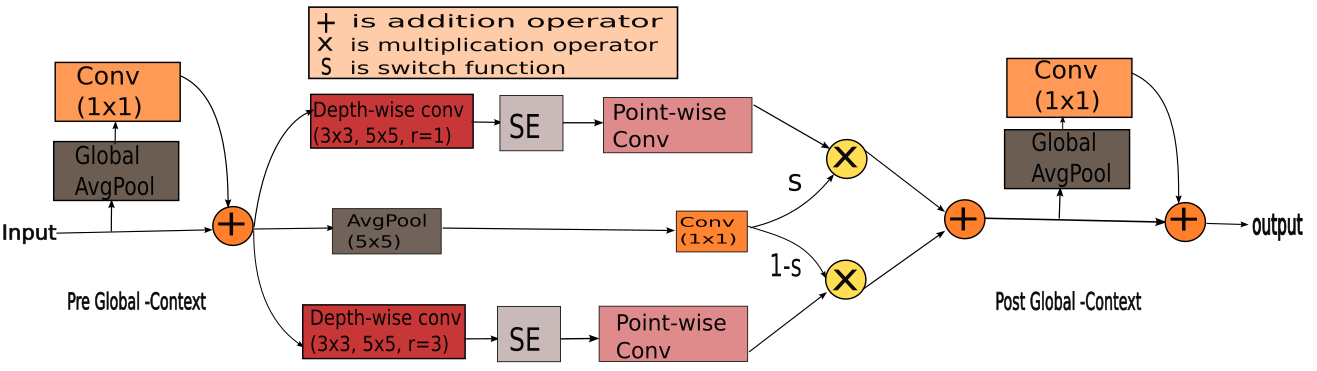}
     \caption{Depthwise atrous  with pointwise switchable Conv layer(DAPSC) . We convert each $3\times3 $ convolutional layer in the baseline Efficientnet to DAC, which gradually alternates the atrous rates used for convolutional computation and added a pointwise switch function(PSC). Two global context modules add image-level information to the features.}
     \label{fig:3}
 \end{figure}

The proposed DAPSC module shifts the switch function towards the pointwise layer instead of the depthwise layers as illustrated in Fig \ref{fig:3}. We apply a depthwise switchable atrous Conv layer with different atrous rates $(DConv(x,w,r))$ and the switch function $S(x)$ after the pointwise layer (PConv(x)) and got the output $PConv(x,w,r)$ (as shown in equation (3)). We use $y = PConv(x, w, r)$ to denote the point-wise convolutional operation with weight $w$ and atrous rate $r=3$ and $x$ as input and output $y$.
\begin{equation}\label{eq4}
PConv(x,w,1) = DConv(x, w,1)\longrightarrow SE(x)\longrightarrow(PConv(x)) 
\end{equation}
\begin{equation}\label{eq5}
PConv(x,w,r) = DConv(x, w,r)\longrightarrow SE(x)\longrightarrow(PConv(x))
\end{equation}
Combining (\ref{eq4}) and (\ref{eq5}) with a switch function $S(x)$ results in the following:
\begin{equation}\label{eq6}
PConv(x, w,1)\longrightarrow S(x) . PConv(x, w,1) + (1-S(x) ) . PConv(x,w,r)
 \end{equation}

In (\ref{eq4}), the input feature map $x$ is first passed through a DConv layer with an atrous rate 1, denoted as $DConv(x, w, 1)$. The output of the DConv layer is then passed through a squeeze-and-excitation (SE) layer, denoted as $SE(x)$, which adaptively recalibrates the feature responses based on channel-wise information. Finally, the output of the SE layer is passed through a point-wise convolutional layer, denoted as $PConv(x)$, with an atrous rate of 1.

Similarly, in  (\ref{eq5}), the input feature map $x$ is passed through a DConv layer with atrous rate 3, denoted as $DConv(x, w, 3)$. The output of the DConv layer is then passed through a SE layer, denoted as $SE(x)$, and finally, through a PConv layer with an atrous rate of 1.

The combination of  (\ref{eq4}) and (\ref{eq5}) with a switch function ($S(x)$) results in the following operation (\ref{eq6}): for each location in the feature map, the switch function generates a binary mask that decides which operation to perform - either pass the feature map through $PConv(x, w, 1)$ or $PConv(x, w, r)$ (with $r=3$). The binary mask is then used to select either the output of $PConv(x, w, 1)$ or the output of $PConv(x, w, r)$ for each location in the feature map, and the selected feature maps are combined using element-wise addition to produce the final output feature map.

We also train the network with and without a global context 
modules whose results are shown in the ablation study.

Next, we illustrate how we incorporate the global context block before and after the proposed depthwise convolution layer.

\subsection{Global context before and after the depthwise convolution layers}\label{subsec2}
\begin{figure}
     \centering
     \includegraphics[height=6cm,width=0.8
    \textwidth]{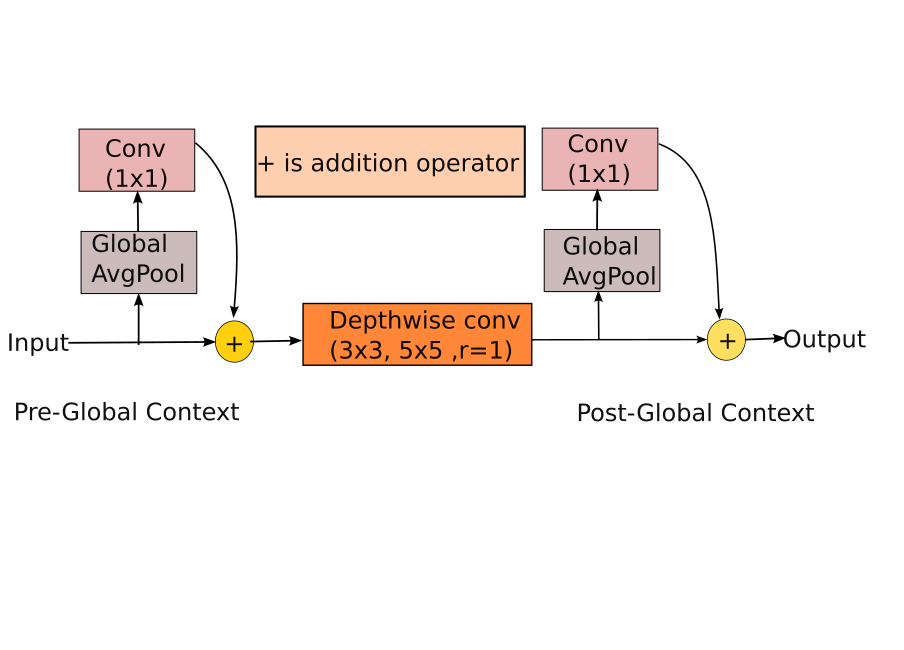}
    \vspace{-2cm}
     \caption{Global context before and after the depthwise convolution layers}
     \label{fig:1}
 \end{figure}

Two global context modules are inserted before and after the depthwise separable Conv layer's primary component as shown in Fig.\ref{fig:1}. These two modules are lightweight because a global average pooling layer first compresses the input characteristics. The results are incorporated back into the mainstream. We observe that the detection performance is improved by including the global context data before the depthwise separable Conv layers $DConv(x,w,1)$, Where, $x$ , $w$ and 1, are input weight and  atrous rate.
The performances of depth-wise separable Conv layers with and without the global context modules are shown in the ablation research.
\begin{equation}\label{eq7}
DConv(x,w,1)
\longrightarrow PrG(x) + DConv(x,w,1) +PoG(x)
\end{equation}
where $PrG(x)$ is Pre-Global Context, $DConv(x,w,1)$ Depthwise separable conv and $PoG(x)$ Post Global Context operator.
Equation (\ref{eq7}) describes the operation of a modified version of the depthwise separable convolutional layer (DConv) that includes pre-global and post-global context operators. The input feature map x is first passed through the DConv layer with an atrous rate of 1, denoted as DConv(x, w, 1) with filter weight $w$. This operation is followed by the Pre-Global Context operator $P rG(x)$, a non-linear transformation applied to the output of the depthwise separable convolution.

The resulting feature map is passed through the post-global context operator, denoted as $PoG(x)$. This operator performs further processing that considers the global context information of the input feature map. The $PoG(x)$ output is then added to the feature map resulting from the previous step, again using element-wise addition.

The combined feature map produced by  (\ref{eq7}) contains local and global context information.

\section{Experiments}\label{sec6}
We first illustrate the experimental setup, followed by a description of the dataset used in the study.

\subsection{Implementation Details}\label{subsec4}
In our implementation, the weights and the biases in the global context modules are initialized with 0. The weight in the switch $S$ is initialized with 0; the bias is set to 1. The initialization method described above ensures that loading the backbone that has been previously trained on EfficientDet and converting all of its $3\times3$ convolutional layers to DSAC will not affect the output before beginning any training on the dataset. The kernel sizes in the depthwise convolution layer of the EfficientDet model are kept as [3, 5], as in \cite{efficientdet}. We use an atrous rate of [3,3] and padding of [3,6] to convert this layer to DSAC, to cope up with the feature dimensions.

\subsection{Experiments}\label{subsec2}
In all of our experiments, we use pre-trained models. During the training process, we use the train2017 set and then used the val2017 set for validation.  We report results on bounding box object detection. We initialize the model using pre-trained weight and train the whole network from scratch until epoch 2/3. We use the PyTorch platform for the experiment. Each model is trained using an SGD optimizer with a momentum of 0.9 and weight decay 4e-5. Focal loss is a modification of cross-entropy loss designed to address the class imbalance problem in object detection tasks.
Focal loss introduces two additional parameters: the focusing parameter $\alpha$ and the modulation parameter $\gamma$. The focusing parameter $\alpha$ is used to down-weight the loss assigned to well-classified examples, to focus more on misclassified complex samples. A common value for $\alpha$ is 0.25.

The modulation parameter $\gamma $ modulates the loss based on the predicted class probability. Specifically, the loss is multiplied by $(1 - p_t)^\gamma$ where $p_t$ is the predicted probability of the true class. This increases the contribution of easy examples to the loss and down-weights the contribution of well-classified hard examples. A common value for $\gamma$ is 1.5.
EfficientDet also uses an anchor-based approach for object detection, where anchor boxes of different aspect ratios cover a range of object shapes. The aspect ratio is set to [1/2, 1, 2] to capture a variety of object shapes.
\subsection{Dataset}\label{subsec2}

We use the Microsoft COCO dataset \cite{lin2014microsoft} to conduct experiments and validate our object detection method. A large image recognition dataset for object detection, segmentation, and captioning tasks is called the Common Objects in Context (COCO) dataset \cite{lin2014microsoft}. It contains over 330,000 images with over 2.5 million object instances labeled across 80 object categories, such as people, animals, vehicles, and household objects.

We use the metric of mAP to validate the proposed method and comparison against the state-of-the-art.

\section{Results and Discussions}
The results of applying the proposed object detector on the MSCOCO dataset \cite{coco}, compared to the state-of-the-art, in terms of mAP percentage, is shown in Table \ref{res}. Clearly, the proposed atrous convolution-based approach alongwith the depthwise convolution scheme outperformed the state-of-the-art by a significant margin.
\begin{table}
\setlength{\arrayrulewidth}{0.2mm}
\setlength{\tabcolsep}{16pt}
\renewcommand{\arraystretch}{1.2}

    \begin{center}
        \caption{Performance (in terms of mAP percentage) of the proposed method with the best performing EfficientNet-d7 backbone, when applied to the MSCOCO dataset, compared to the state-of-the-art.}\label{res}
        \begin{tabular}{c|c|c} \hline
            Method & Published & mAP\% \\ \hline
            EfficientDet \cite{efficientdet} & CVPR 20 & 50.15 \\ \hline
            DetectorS \cite{detectors} & CVPR 21 & 50.00 \\ \hline
            Li et al. \cite{paa2} & Pattern Analysis and App. 22 & 48.58 \\ \hline
            Liu et al. \cite{paa1} & Pattern Analysis and App. 23 & 48.81 \\ \hline
            Jiang et al. \cite{icme22} & ICME 22 & 48.70 \\ \hline
            Xu et al. \cite{aploss} & CVPR 22 & 50.00 \\ \hline
            Zheng et al. \cite{distillation} & CVPR 22 & 50.20 \\ \hline
            Gilg et al. \cite{gilg_wacv23} & WACV 23 & 49.78 \\ \hline
            Proposed Method & ------- & 51.32 \\ \hline
        \end{tabular}
    \end{center}
\end{table}

\subsection{Ablation Studies}\label{sec4}
We conduct several ablation studies on the proposed method, to experiment it's efficacy. Results of our ablation studies are illustrated in Table \ref{tab2}. We start experimenting with the Efficientnet-d0, d1, and d2 backbones without any modification. Next, we apply the proposed global context on all three backbones and observed around 1\% improvement in the mAP measure. We further applied the proposed Depthwise Switchable Atrous Convolution (DSAC) scheme, to observe a little increase in the mAP measure, for all three backbones. Next, we applied the proposed Depthwise Atrous with Pointwise Switchable Convolution (DAPSC), alongwith the global context, which provides the highest value of mAP measure. Hence, Table \ref{tab2} shows the importance of all three major contributions made in this study.

\begin{table}[h]
\begin{center}
\caption{Ablation studies with the proposed method when applied to the MSCOCO dataset. We gradually applied the Global context, DSAC, and DAPSC schemes, and observed how each of these contributions helped the proposed method to achieve state-of-the-art accuracy. Here we have shown the results of applying only the Efficientnet d0, d1, and d2 as backbones.}\label{tab2}
\begin{tabular*}{\textwidth}{@{} l @{\extracolsep{\fill}} *{14}{c} @{}}
\toprule%
 Model & Epoch & Learning Rate & Optimizer & Batch Size & mAP \\
\midrule
D0(Original)  & 1 & 0.00001 & SGD & 16 & 31.5 \\
D1(Original)  & 1 & 0.00001  & SGD  & 8 & 36.3 \\
D2(Original)  & 1 & 0.00001  & SGD  & 4  & 39.0 \\
D0+Global context & 2 & 0.0001  & SGD  & 16 & 32.2 \\
D1+Global context  & 2 &  0.0001 & SGD  & 8 & 37.8 \\
D2+Global context & 2 & 0.0001 & SGD & 4 & 40.7 \\
D0+DSAC & 3 & 0.00001  & SGd  & 8 & 31.6 \\
D1+DSAC  & 3 & 0.00001  & SGD  & 8 & 38. 0 \\
D2+DSAC  & 3 & 0.00001  & SGD  & 4 & 40.6 \\
D0+DSAC+Global Context & 2 & 0.00001  & SGD  & 16 & 32.5 \\
D1+DSAC+Global Context  & 2 & 0.00001  & SGD  & 4 & 38.8 \\
D2+DSAC+Global Context & 2 & 0.00001  & SGD  & 4 & 40.7 \\
D0+DAPSC & 2 & 0.00001  & SGD  & 16 & 32.3 \\
D1+DAPSC & 2 & 0.00001  & SGD  & 4 & 38.1 \\
D2+DAPSC & 2 & 0.00001  & SGD  & 4 & 41.0 \\
D0+DAPSC+ Global Context & 2 & 0.00001  & SGD  & 16 & 32.4 \\
D1+DAPSC+ Global Context & 2 & 0.00001  & SGD  & 4 & 37.5 \\
D2+DAPSC+ Global Context & 2 & 0.00001  & SGD  & 4 & 40.6 \\
\botrule
\end{tabular*}

\end{center}
\end{table}

\subsection{Discussion}\label{sec12}
From Tables \ref{res} and \ref{tab2}, we observe that the proposed global context enhances the efficacy of the proposed method significantly, due to the scale-invariant nature. The proposed atrous convolution scheme on the EfficientDet backbone has a mild effect on the efficacy, however, this scheme can reduce the parameter count drastically. We also observe that, the proposed DAPSC scheme improves the efficacy by a significant margin.

\section{Conclusion}\label{sec13}
A depthwise switchable atrous convolutional network is proposed in this research work. The proposed model equipped a switchable mechanism to control the use of different rates of atrous (dilated) convolution operations, including traditional atrous convolution, depthwise atrous convolution, and pointwise atrous convolution. The idea behind the depthwise switchable atrous convolutional network is to allow the network to automatically switch between these different types of atrous convolution operations based on the input data so that it can be useful for detecting objects that appear in different scales in the images, and improve the accuracy of the proposed method. The proposed scale-invariant feature can be extended to work on videos, to track objects across frames with varying scales. Further, the proposed atrous convolution can be tested with the YOLO family of object detectors, to analyze the effect.

\section{Declaration}

$\mathbf{Funding:}$ The authors did not receive support from any organization for the submitted work.

$\mathbf{Competing~Interest:}$ The authors have no competing interests to declare that are relevant to the content of this article.

$\mathbf{Compliance~with~ethical~standards:}$ The authors have no conflict of interest to disclose. Further, the authors certify that, the research presented in this article does not involve any human participants or animals.

$\mathbf{Data~availability~statement:}$ We declare that, this study does not contain any data.

$\mathbf{Author~Contribution~statement:}$ Amrita Singh did all the experiments, coding, and wrote the first draft of the paper. Snehasis Mukherjee was involved in supervising, analyzing the result, ideation, and writing the final draft of the paper.

\bibliography{sn-bibliography}

\end{document}